\pdfoutput=1

\documentclass[11pt]{article}

\usepackage[]{eacl2023}

\usepackage{times}
\usepackage{latexsym}
\usepackage{booktabs}
\usepackage{amsmath}
\usepackage[T1]{fontenc}

\usepackage[T2A, T1]{fontenc}
\usepackage[utf8]{inputenc}
\usepackage[russian, english]{babel}

\usepackage{microtype}
\usepackage{graphicx}
\usepackage{enumitem}
\usepackage{colortbl}
\usepackage{multirow}

\title{Multilingual Bidirectional Unsupervised Translation Through Multilingual Finetuning and Back-Translation}

\author{Bryan Li$^1$\Thanks{ Correspondence to: \texttt{bryanli@seas.upenn.edu}}
, Mohammad Sadegh Rasooli$^2$, Ajay Patel$^1$, Chris Callison-Burch$^1$ \\
  $^1$University of Pennsylvania, Philadelphia, PA, USA \\ $^2$Microsoft, Mountain View, CA, USA
  }
\begin{document}
\maketitle

\begin{abstract}
We propose a two-stage approach for training a single NMT model to translate unseen languages both to and from English.
For the first stage, we initialize an encoder-decoder model to pretrained XLM-R and RoBERTa weights, then perform multilingual fine-tuning on parallel data in 40 languages to English. We find this model can generalize to zero-shot translations on unseen languages. For the second stage, we leverage this generalization ability to generate synthetic parallel data from monolingual datasets, then bidirectionally train with successive rounds of back-translation.

Our approach, which we  EcXTra (\underline{E}nglish-\underline{c}entric Crosslingual (\underline{X}) \underline{Tra}nsfer), is conceptually simple, only using a standard cross-entropy objective throughout. It is also data-driven, sequentially leveraging auxiliary parallel data and monolingual data. We evaluate unsupervised NMT results for 7 low-resource languages, and find that each round of back-translation training further refines bidirectional performance. Our final single EcXTra-trained model achieves competitive translation performance in all translation directions, notably establishing a new state-of-the-art for English-to-Kazakh (22.9 > 10.4 BLEU).
\end{abstract}


\newcolumntype{g}{>{\columncolor[gray]{0.9}}c}

\newcommand{\ecresults}{
    \begin{table*}[htb!]
    \centering
    \setlength{\tabcolsep}{4pt}
    \begin{tabular}{@{}l|cgcgcgcgcgcgcgcg}
    \toprule
      \multirow{2}{*}{Round} & \multicolumn{2}{c}{kk-en} & \multicolumn{2}{c}{gu-en} & \multicolumn{2}{c}{si-en} & \multicolumn{2}{c}{ne-en} & \multicolumn{2}{c}{ps-en} & \multicolumn{2}{c}{is-en} & \multicolumn{2}{c}{my-en} & \multicolumn{2}{c}{Avg.} \\ 
    & $\rightarrow$ & $\leftarrow$ & $\rightarrow$ & $\leftarrow$ & $\rightarrow$ & $\leftarrow$ & $\rightarrow$ & $\leftarrow$ & $\rightarrow$ & $\leftarrow$ & $\rightarrow$ & $\leftarrow$ & $\rightarrow$ & $\leftarrow$ & $\rightarrow$ & $\leftarrow$ \\
     \midrule
    $r_0$ & 19.6 & n/a & 23.2 & n/a & 17.5 & n/a & 20.9 & n/a & 9.8 & n/a & 26.0 & n/a & 16.5 & n/a & 19.1 & n/a \\
     \midrule
    $r_1$ & \textbf{18.5} & 20.7 & 21.1 & 13.1 & 14.8 & 6.6 & 18.0 & 8.3  & 9.0 & 8.0 & 24.4 & 23.4 & \textbf{14.3} & 8.3 & 17.2 & 12.6 \\
    $r_2$ & 18.2 & \textbf{22.9} & \textbf{21.5} & \textbf{13.9} & \textbf{17.8} & \textbf{7.1} & \textbf{19.7} & \textbf{9.3} & \textbf{13.0} & \textbf{8.1} & \textbf{30.6} & \textbf{25.4} & 12.9 & \textbf{8.8} & \textbf{19.1} & \textbf{13.6} \\
    \bottomrule
    \end{tabular}
    \caption{BLEU scores for various rounds of EcXTra models on several low-resource translation test sets. The row divisions indicate groups by approach: zero-shot (no synthetic parallel data), unsupervised (synthetic parallel data). Foreign-English translation ($\rightarrow$) columns are in white, while English-foreign ($\leftarrow$) columns are in grey. `Avg.' is the unweighted average BLEU scores across that translation direction. `n/a' indicates unsupported directions.
    For the second group, the best BLEU score per column is \textbf{bolded}.}
    \label{tab:ecresults}
    \end{table*}
}

\newcommand{\cmpresults}{
    \begin{table*}[htb!]
    \centering
    \setlength{\tabcolsep}{4pt}
    \begin{tabular}{@{}l|cgcgcgcgcgcgcg}
    \toprule
     \multirow{2}{*}{Round} & \multicolumn{2}{c}{kk-en} & \multicolumn{2}{c}{gu-en} & \multicolumn{2}{c}{si-en} & \multicolumn{2}{c}{ne-en} & \multicolumn{2}{c}{ps-en} & \multicolumn{2}{c}{is-en} & \multicolumn{2}{c}{my-en} \\ 
    & $\rightarrow$ & $\leftarrow$ & $\rightarrow$ & $\leftarrow$ & $\rightarrow$ & $\leftarrow$ & $\rightarrow$ & $\leftarrow$ & $\rightarrow$ & $\leftarrow$ & $\rightarrow$ & $\leftarrow$ & $\rightarrow$ & $\leftarrow$  \\
     \midrule
    mBART-ft & 19.6 & n/a & 17.3 & n/a & 12.2 & n/a & 14.4 & n/a & 0.9 & n/a & ... & n/a & 3.6 & n/a   \\
    SixT & 19.0 & n/a & 17.3 & n/a & 12.2 & n/a & 14.4 & n/a & 11.4 & n/a & ... & n/a & 5.4 & n/a \\ 
    SixT+ & \textbf{27.3} & n/a & \textbf{27.5} & n/a & \textbf{17.5} & n/a & \textbf{23.7} & n/a & \textbf{12.9} & n/a & ... & n/a & 15.3 & n/a   \\
    EcXTra-$r_0$ & 19.6 & n/a & 23.2 & n/a & \textbf{17.5} & n/a & 20.9 & n/a & 9.8 & n/a & 26.0 & n/a & \textbf{16.5} & n/a \\
    \midrule
    \citet{garcia-etal-2021-harnessing} & 16.4 & 10.4 & \textbf{22.2} & \textbf{16.4} & 16.2 & \textbf{7.9} & \textbf{21.7} & 8.9 & n/a & n/a & n/a & n/a & n/a & n/a \\
    EcXTra-$r_2$ & \textbf{18.2} & \textbf{22.9} & 21.5 & 13.9 & \textbf{17.8} & 7.1 & 19.7 & \textbf{9.3} & 13.0 & 8.1 & 30.6 & 25.4 & 12.9 & 8.8  \\
    \midrule
    Supervised$^{1234567}$ & ... & 12.1 & ... & 28.2 & ... & 6.5 & ... & 26.3 & ... & 11.0 & ... & 23.6 & ... & 13.9 \\
    \bottomrule
    \end{tabular}
    \caption{
        BLEU scores comparing various models to EcXTra. The row divisions indicate groups by approach: zero-shot (no synthetic parallel data), unsupervised (synthetic parallel data), and supervised (real parallel data). `n/a' indicates unsupported directions, while `...' indicates results not provided. Within a row group, the best BLEU score per column is \textbf{bolded}.
        Supervised results, from left to right: $^1$\citet{rasooli-etal-2021-wikily} $^2$\citet{li-etal-2019-niutrans}  $^3$\citet{bei-etal-2019-gtcom} $^4$\citet{ko-etal-2021-adapting} $^5$\citet{shi-etal-2020-oppos} $^6$\citet{simonarson-etal-2021-mideinds} $^7$\citet{hlaing-etal-2021-nectecs} 
    }
    \label{tab:cmpresults}
    \end{table*}
}

\newcommand{\mtoecheck}{
    \begin{table}[tb!]
    \centering
    \setlength{\tabcolsep}{4pt}
    \begin{tabular}{@{}l|ccccc}
    \toprule
      Round & zh-en & hi-en & tr-en & ru-en & Avg. \\
     \midrule
    $r_0$  & 19.2 & 21.9 & 28.5 & 34.0 & 25.9 \\
    $r_1$  & 17.0 & 17.6 & 26.2 & 32.5 & 23.3 \\
    $r_2$  & 17.4 & 16.0 & 27.1  & 32.9 & 23.3 \\
    \bottomrule
    \end{tabular}
    \caption{BLEU scores for each EcXtra training round on several supervised foreign-English translations.}
    \label{tab:m2echeck}
    \end{table}
}

\section{Introduction}
Current neural machine translation (NMT) systems owe much of their success to efficient training over large corpora of parallel sentences, and consequently tend to struggle in low-resource scenarios and domains~\cite{kimWhenWhyUnsupervised2020,marchisio-etal-2020-unsupervised}. This has motivated investigation into the field of zero-resource NMT, in which no parallel sentences are available for the source-target language pair. This is especially valuable for low-resource languages, which by nature have little to no parallel data.

There are two mainstream lines of inquiry towards developing models to tackle zero-resource machine translation. \emph{Unsupervised machine translation} learns a model from monolingual data from the source and target languages. Some research involves introducing new unsupervised pre-training objectives between monolingual datasets~\cite{lample2019cross, artetxe2019effective}. Others devise training schemes with composite loss functions on various objectives~\cite{ko-etal-2021-adapting, garcia-etal-2021-harnessing}. In contrast, \emph{zero-shot machine translation} learns a model by training on other datasets~\cite{liu-etal-2020-multilingual-denoising} or other language pairs~\cite{chen-etal-2021-zero,chenMakingMostMultilingual2022}, then directly employ this model for translating unseen languages.

This work leverages both mainstream approaches in zero-resource translation. 
We propose a conceptually simple, yet effective, two-stage approach for training a single NMT  model to translate unseen languages both to and from English. The first stage model is trained on \textit{real} parallel data from 40 high-resource languages to English. This results in a strong zero-shot model, which we use to translate unseen languages to English. By applying back-translation to flip the order, we obtain English-to-unseen \textit{synthetic} parallel data. In the second stage, we continue training the model on successive rounds of offline back-translation, where each round uses the prior round for both for weight initialization and for synthetic parallel data.

We term our overall unsupervised translation approach EcXTra (\underline{E}nglish-\underline{c}entric Crosslingual (\underline{X}) \underline{Tra}nsfer).
EcXTra can be thought of as a data-driven approach, which sequentially leverages auxiliary parallel data then monolingual data. Each stage's model is initialized to an informed pretrained model, before fine-tuning. We initialize the first stage model's encoder and decoder to XLM-RoBERTa~\cite{conneau2020unsupervised} and RoBERTa~\cite{liuRoBERTaRobustlyOptimized2019} respectively, and we initialize the second stage model's weights to those of the first stage. In doing so, EcXTRa importantly avoids the complicated training schemes and custom training objectives of prior work. 

As our approach is simple to train and extend to new unseen languages, we release all code, data and pretrained models.\footnote{\url{https://github.com/manestay/EcXTra}} Our contributions are:
\begin{enumerate}
    \item We introduce EcXTra, a two-stage approach for training a single NMT model to translate unseen languages to and from English. In its two stages, EcXTra combines zero-shot NMT and unsupervised NMT: multilingual fine-tuning and back-translation respectively.
    \item  Our work is an empirical study of an agnostic view towards multilinguality, as we train the zero-shot stage on balanced splits of parallel data from 40 languages to English. In contrast, prior work has largely explored multilinguality by selecting train languages with oracle knowledge of the test languages.
    \item We evaluate the bidirectional unsupervised NMT performance of a single EcXTra-trained model on 7 foreign-English test sets (14 total). This final model, trained in two rounds of back-translation, achieves competitive unsupervised performance for most language directions, establishing a new state-of-the-art for English-Kazakh. We are also the first to report, the best of our knowledge, unsupervised results for 3 translation directions: English-Pashto, English-Myanmar, and English-Icelandic.
\end{enumerate}

\section{Our Approach}
\label{sec:approach}

Our training procedure closely follows the standard machine translation task. \textit{Machine translation} involves developing models to output text in a target language $\mathcal{T}$, given text in a source language $\mathcal{S}$. In a typical supervised MT setting, it is assumed there is a parallel corpus ${\cal P}=\{(s_i,t_i)\}_{i=1}^{n}$ in which each sentence $t_i \in \mathcal{T}$ is a translation of $s_i \in \mathcal{S}$. A model is then trained on these examples, to minimize the cross-entropy loss given by
\begin{equation}\label{eq:1}
    {\cal L}({\cal P}; \theta) = \sum_{i=1}^{n} \log p(t_{i} | s_i; \theta)
\end{equation}
where $\theta$ is a collection of learned parameters.

Given enough parallel data, this training framework allows contemporary NMT models to achieve strong performance~\cite{dabreSurveyMultilingualNeural2020}. However, in the unsupervised setting arises the fundamental challenge that we no longer have any parallel data between the source and target languages of interest.

Conceptually, we divide the two stages of our training procedure into four steps: 
\paragraph{1a.} \textit{Zero-shot model transfer} by initializing to pretrained multilingual LMs. We use  an XLM-RoBERTa encoder and a RoBERTa decoder.
\paragraph{1b.} \textit{Multilingual fine-tuning} for this initialized model, on parallel data from diverse source languages to English.
\paragraph{2a.} \textit{Synthetic parallel data creation} using back-translations from the stage 1 model.
\paragraph{2b.} \textit{Back-translation training} by initializing to the stage 1 model, then further training on the synthetic parallel data, in both translation directions. Steps 2a and 2b are iterated for several rounds, in each initializing to the prior round model.

Observe that these are are widely-used techniques in the field of machine translation. Our main contribution is in presenting an effective synthesis of the techniques to enable a single model to perform zero-shot and bidirectional translation (while using only a standard loss function).


\paragraph{Terminology}
It is worthwhile formalizing our exact terminology, given that prior work in this field uses terms rather inconsistently.\footnote{See Section 2.1 of \citet{garcia-etal-2021-harnessing} for further discussion on this inconsistency.}
Our setting is \textit{English-centric}, as the language pairs include English as either the source or target\footnote{We focus on the English-centric setting because it is the language with the most parallel data to other languages.} Our final model is \textit{bidirectional}, in that it can translate $\mathcal{S}$ to $\mathcal{T}$ and also translate $\mathcal{T}$ to $\mathcal{S}$. We call the non-English side of a pair a \textit{foreign} language. Therefore, we use the terms foreign-English and many-to-English interchangeably (likewise with English-foreign and English-to-any).  Languages seen during training on parallel datasets are \textit{auxiliary} languages.

\subsection{Zero-shot Model Transfer}
There are many structural as well as lexical similarities across different languages, especially within language families.
By training a multilingual translation model on gold-standard parallel datasets for auxiliary higher-resource languages, we aim to exploit these similarities. Specifically, we train model parameters $\theta$ on parallel data between $n$ auxiliary languages $\mathcal{S} = \mathcal{S}_1 \ldots\mathcal{S}_n$ and some target language $\mathcal{T}$ (for us, English). The goal is to have the model learn to generalize to translating $m$ unseen language data $\mathcal{U} = \mathcal{U}_1 \ldots\mathcal{U}_m$  to $\mathcal{T}$. In other words, in the absence of gold-standard parallel data ${\cal P}$ in our zero-resource languages, we make use of knowledge transfer from larger parallel datasets with auxiliary source languages.  Looking back at Equation~\ref{eq:1}, we redefine its objective function as 
\begin{equation}
\sum_{i=1}^{n}  {\cal L}(D({\cal S}_i, {\cal T}); \theta) 
\end{equation}
where $D({\cal S}_i, {\cal T})$ is the gold-standard parallel dataset for language ${\cal S}_i$ and English (${\cal T}$).

\paragraph{EcXTRA: Multilingual fine-tuning}
Multilinguality, namely having diverse auxiliary languages
is key to good zero-resource NMT performance~\cite{garcia-etal-2021-harnessing}. In this setting, because there are no true $(s_i,t_i)$ examples until inference time, performance becomes especially sensitive to the initialization of parameters $\theta$. We do so by initializing the encoder with XLM-RoBERTa and decoder with RoBERTa. The former allows for transfer learning from strong pretrained models that are already trained on monolingual data in languages (including the unseen languages of interest), whereas the latter allows for a good understanding of fluent English sentences. Initializing the encoder and decoder to pretrained LMs follows prior work \cite{rothe2020leveraging, ma2020xlm}.

From this initialization, we then fine-tune the model on parallel data from many high-resource languages to English. The resulting model is able to translate from unseen language to English, but not the other way. We next discuss how we extend our approach to develop a bidirectional model.

\subsection{Synthetic Parallel Data Creation}
We assume in this step that we have monolingual data in the unseen languages, which are typically collected by crawling web data. We make use of the model trained in the previous stage to translate all the monolingual sentences $(s_j)_{j=1}^{k}$ to English, thereby having synthetic parallel data $(s_j, \hat{t}_j)_{j=1}^{k}$ where $\hat{t}_i$ is the translation output from the zero-shot model. We then flip the order in each pair to produce examples $\hat{\cal P} = (\hat{t}_j, s_j)_{j=1}^{k}$, then continue training. This process of bootstrapping additional data is called (offline) \textit{back-translation}.

While back-translation is typically used in low-resource settings, our approach extends it towards the zero-resource setting.
We perform back-translation for all unseen languages, and concatenating together all synthetic parallel data $(\hat{\cal P}_i)_{i=1}^{m}$.

\paragraph{EcXTRA: Training on Synthetic Data}
In this step, we train a bidirectional English-centric model. We ensure bidirectionality by training on both the English-foreign synthetic parallel data, and the foreign-English auxiliary parallel data. Our new objective function is thus a combination of the two cross-entropy losses:
\[
\sum_{i=1}^{n}  {\cal L}(D({\cal S}_i, {\cal T}); \theta) + \sum_{i=1}^{m}  {\cal L}(\hat{\cal P}_i; \theta)
\]

Just as we initialized the zero-shot model to pretrained multilingual LMs, so too do we initialize the unsupervised model to the zero-shot model.
After training an initial unsupervised bidirectional model, we further refine performance by running iterative rounds of the synthetic parallel data creation and training process. 

\section{Datasets Used}
\label{sec:data}
Here we succinctly describe the data, providing further details in Appendix~\ref{appsec:data}. 

\paragraph{Training}
For the zero-shot stage, we use parallel corpora from higher-resource auxiliary languages to English. 
We utilize a subset of the Many-to-English v1 dataset~\cite{gowda-etal-2021-many}.
We consider only the 40 largest  foreign-English pairs,\footnote{Codes for training languages (with those used for validation in bold): \textbf{tr}, sr, \textbf{fr}, he, \textbf{ru}, ar, \textbf{zh}, bs, nl, \textbf{de}, pt, no, \textbf{it}, \textbf{es}, pl, \textbf{fi}, fa, sv, da, el, \textbf{hu}, sl, vi, \textbf{et}, sk, ja, \textbf{lt}, \textbf{lv}, uk, th, \textbf{cs}, ko, id, ca, mt, \textbf{ro}, bg, hr, \textbf{hi}, eu} and equally sample 2 million examples from each.\footnote{The rationale is further discussed in Section~\ref{sec:limitations}.}

The resulting dataset, which we term \textit{m2e-40}, consists of 80 million sentence pairs from 40 source languages. Note that unlike most prior work, we have taken an agnostic view towards multilinguality --- we do not choose the training languages with reference to the testing languages.

For the unsupervised stage, we use monolingual corpora in the 7 test languages (below) from CommonCrawl and CC-100.

\paragraph{Testing}
We evaluate our approach on 7 languages: Kazakh (kk), Gujarati (gu), Sinhala (si), Nepali (ne), Pashto (ps), Icelandic (is), and Burmese (my). Test sets are taken from WMT21, FLORES-101 and WAT21. The languages were chosen for both their diversity and for comparison to prior unsupervised NMT work. 

\paragraph{Validation}
To validate the zero-shot stage, we select 15 foreign-English parallel datasets from WMT19 development data; these languages are seen during training.

In the unsupervised stage we only have access to monolingual data. For validation purposes, we thus reserve a small number of synthetic sentence pairs (250 per direction * 14 directions).

\begin{figure}[t!]
    \centering
    \includegraphics[width=.4\textwidth]{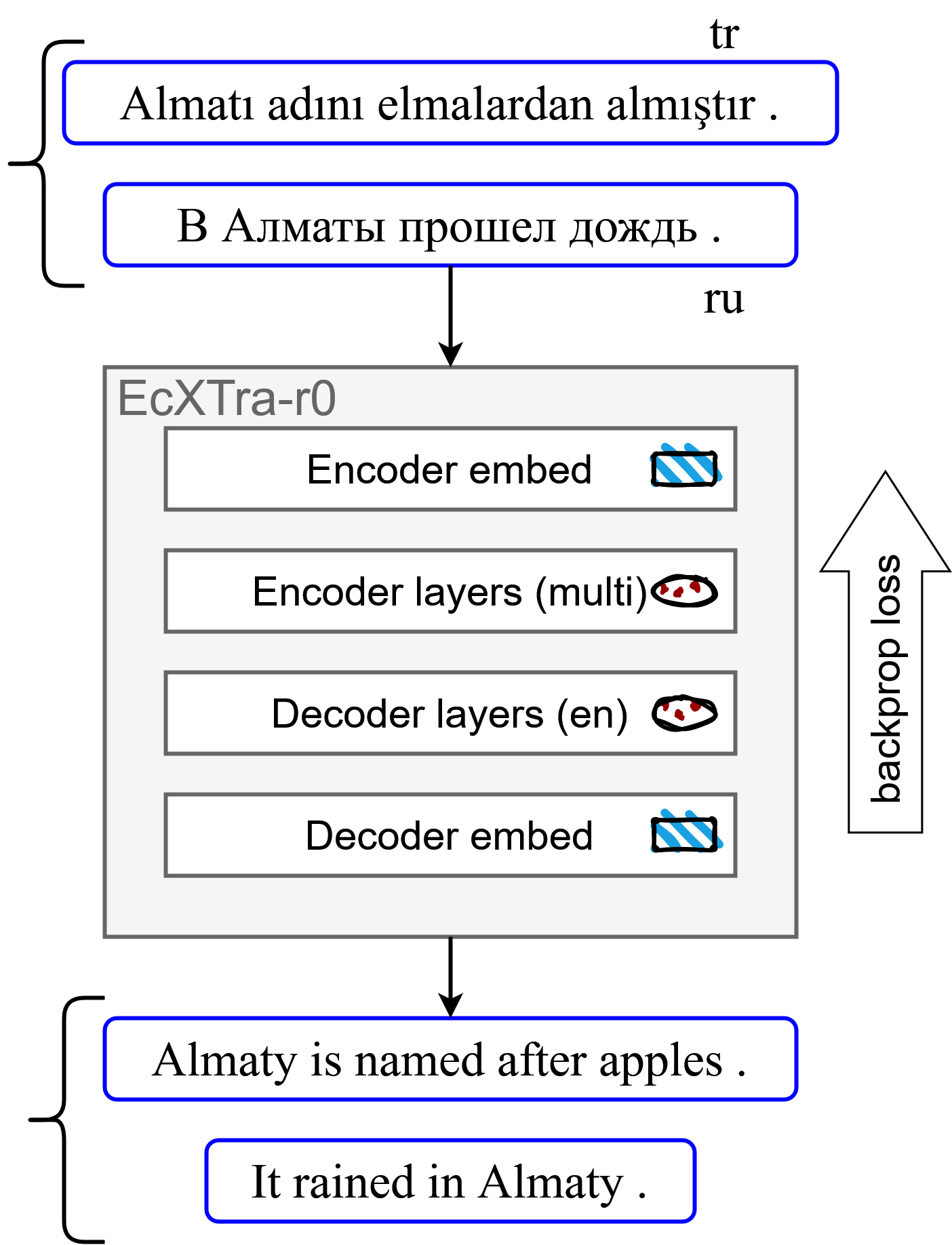}
    \caption{An illustration of the first stage of training (or EcXTra-$r_0$). The model learns to translate foreign sentences to English. The encoder is initialized to XLM-RoBERTa, and the decoder is initialized to RoBERTa. Both embeddings are frozen (blue rectangle), while layers are finetuned (red ellipse).}
    \label{fig:ecxtra}
\end{figure}

\begin{figure*}[ht!]
    \centering
    \includegraphics[width=.75\textwidth]{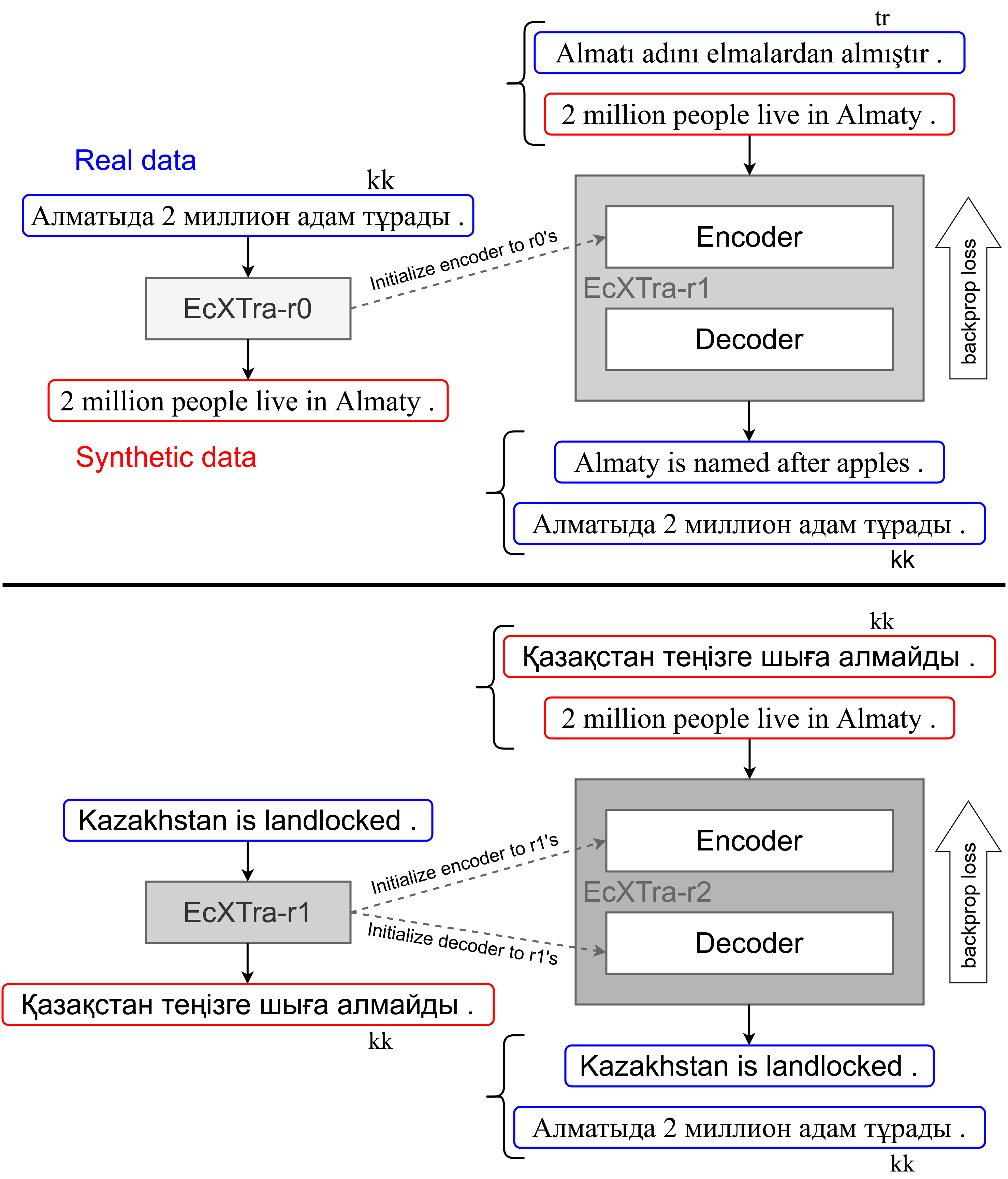}
    \caption{An illustration of the second stage of training, split into 2 rounds. Each round $n$ is trained on a concatenation of back-translations from round $n-1$, and the the opposite direction training data from round $n-1$. Round 1 uses English-foreign synthetic data and transfers only the encoder, while round 2 uses synthetic data for both directions and transfers both encoder and decoder. Note that EcXTra blocks are abbreviated from Figure~\ref{fig:ecxtra}.}
    \label{fig:bt}
\end{figure*}

\section{Experimental Setup}
We move from the overall EcXTra approach, to the specifics of using EcXTra to train an NMT model.

\subsection{Stage 1: Multilingual Fine-Tuning}
\textit{Multilingual fine-tuning} is the process of training a many-to-English zero-shot NMT model on parallel data from auxiliary languages to English. Figure~\ref{fig:ecxtra} depicts the multilingual fine-tuning process.

\paragraph{Architecture}
We use an encoder-decoder, Transformer-based NMT model.
Encoder layers and embeddings are initialized to XLM-R large, and decoder layers and embeddings are initialized to RoBERTa-large. These models were pretrained on a large multilingual corpora with various self-supervised language objectives.
The encoder vocabulary is from XLM-R, and the decoder vocabulary is from RoBERTa.

\paragraph{Setup}
In the multilingual fine-tuning stage, we fine-tune our initialized model on WikiMatrix-25en. We freeze both the encoder and decoder embeddings and fine-tune both the encoder and decoder layers. This model thus has 0.76B trainable parameters (1.1B total). We select the best model checkpoint using early stopping. 

Our training scheme uses the same supervised training objective of standard supervised NMT models. We hypothesize that this training scheme unlocks the cross-lingual transferability of XLM-R to zero-shot settings, with the same reasoning as~\citet{chenMakingMostMultilingual2022}.

\subsection{Stage 2: Back-Translation}
In the unsupervised stage, we perform offline back-translation to bootstrap from foreign-English translation to English-foreign (and back).  Figure~\ref{fig:bt} depicts the back-translation and training process.


\paragraph{Architecture} Most of the architecture is transferred directly from the stage 1 model: encoder embeddings, encoder layers, and decoder layers. We cannot transfer the decoder embeddings, since the model now needs to output multiple languages. Instead, the decoder embeddings are tied to the encoder embeddings, which are frozen XLM-R embeddings. The resulting model thus has 0.96B trainable parameters (1.2B total parameters).

\paragraph{Notation} Recall the zero-shot stage can be thought of as a pre-training step for the unsupervised stage. We thus designate the zero-shot model as EcXTra-$r_0$, and the unsupervised models as EcXTra-$r_i$, where $i$ denotes the current round of back-translation (or simply $r_i$ for brevity). We denote the \textit{m2e-40} dataset as $\mathcal{D}_0$, the concatenation of all foreign monolingual corpora as $\mathcal{D}_{(l)}$, and the English monolingual corpus as $\mathcal{D}_{(e)}$. Synthetic parallel data are $\hat{\mathcal{D}}_{(l)\leftarrow(e)_i}$ or $\hat{\mathcal{D}}_{(e)\leftarrow(l)_i}$.

\paragraph{Training Data}
As 25M parallel sentences were used to train $r_0$, we generate about the same amount (3M per language * 8 languages = 24M) of back-translation data. Each $r_i$ therefore is trained on \textasciitilde 50M sentences, given the bidirectional training.

For each source language sentence, we add a special start token to indicate the desired target language, following the trick of \citet{johnsonGoogleMultilingualNeural2017}. An example is \texttt{<2kk>} to target Kazakh.\footnote{Our specific implementation is detailed in Appendix~\ref{ssec:start_tok}.}

\paragraph{Setup}

Back-translation proceeds in successive stages. The main idea is that, for the current round $r_i$, we use $r_{i-1}$ to generate synthetic parallel data by translating the monolingual corpus---$\mathcal{D}_{(l)}$ for odd rounds, $\mathcal{D}_{(e)}$ for even rounds. The source and target directions are then flipped before being used as training data. We also use $r_{i-1}$ to intialize weights for $r_{i}$. 

In our approach we aim to train bidirectional models. Therefore, the training data of $r_{i}$ consists of both back-translations from $r_{i-1}$, as well as the opposite direction training data used for $r_{i-1}$ itself. Thus the training data for round 1 is $\hat{\mathcal{D}}_{(l)\leftarrow(e)_1} + \mathcal{D}_0$, and for round 2 is $\hat{\mathcal{D}}_{(e)\leftarrow(l)_2} + \hat{\mathcal{D}}_{(l)\leftarrow(e)_1}$.

We ensure that for synthetic parallel data, the target side is always fluent monolingual text. As observed by ~\citet{niu2018bi}, this avoids the possible degradation from training to produce MT output.

For our experiments, we set $m=2$, performing two rounds of back-translation -- consistent with prior findings that improvement tapers off after two rounds \cite{hoang2018iterative}. The final model, EcXTra-$r_2$, will have learned from back-translated data in both directions.

\section{Results}
\ecresults
We evaluate our models on test sets for 7 low-resource-to-English pairs in both translation directions (14 directions total). We use evaluation metrics with are consistent with prior work. By default, we report detokenized sacreBLEU~\cite{post-2018-call}.\footnote{\texttt{BLEU|nrefs:1|case:mixed|eff:no|tok:\\13a|smooth:exp|version:2.0.0}} For the Indic languages (gu, si, ne), we report tokenized BLEU with the Indic-NLP library~\cite{kunchukuttan2020indicnlp}. For Burmese (my), we report SPM-BLEU~\cite{goyal2022flores} to handle the language's optional spacing.

\subsection{Main Results}
\label{ssec:main_results}

Table~\ref{tab:ecresults} shows results for each EcXTra round.

\paragraph{Foreign-English Results ($\rightarrow$)} EcXTra-$r_0$ (or $r_0$) is indeed able to perform zero-shot foreign-English translations. 
The unsupervised $r_1$ has lower scores, this is likely because this model is now tasked with performing 7 additional tasks on top of the original many-to-English task. $r_2$ recovers the overall performance, with the same average BLEU as $r_0$. While $r_2$ underperforms $r_1$ for a few individual pairs, it handily beats $r_0$ for ps-en (13.0 > 9.8) and for is-en (30.6 > 26.0), underscoring the overall quality of the back-translations.

\paragraph{English-Foreign Results ($\leftarrow$)}
Similarly for English-foreign, we observe that $r_2$ matches or exceeds $r_1$ overall across language pairs (13.6 > 12.6). This is in spite of $r_1$ and $r_2$ sharing the same English-foreign training data $\mathcal{D}_{(l)\leftarrow(e)_1}$.

\subsection{Comparisons with Prior Work}
\label{ssec:compare_results}
\cmpresults

Table~\ref{tab:cmpresults} compares the best EcXtra-trained model, $r_2$, with prior work (as well as the zero-shot $r_0$).\footnote{Confidence intervals for our results are not shown, but fall between $\pm 0.4$ to $\pm 1.0$.} We emphasize that these results are \textit{not fully comparable}, given the differing training datasets, models, and number of languages supported.\footnote{More discussion can be found in Section~\ref{sec:limitations}.} However, the comparisons can still illustrate the effectiveness of the language-agnostic nature and simplicity of EcXTra. We compare to:
\paragraph{SixT}~\cite{chen-etal-2021-zero}: trained on a German-English parallel dataset.
\paragraph{SixT+}~\cite{chenMakingMostMultilingual2022}: trained on AUX6, a parallel dataset in 6 high-resource languages. This is concurrent to our work.
\paragraph{mBART-ft}~\cite{tang-etal-2021-multilingual}: mBART-ft is an mBART model further fine-tuned on AUX6.
\paragraph{\citet{garcia-etal-2021-harnessing}}: a single bidirectional unsupervised NMT model trained in 3 stages using combinations of various training objectives on parallel data, real and synthetic (from back-translation).

\paragraph{Zero-Shot NMT Results}
Considering the first four rows of Table~\ref{tab:cmpresults} we see that EcXTra-$r_0$ outperforms mBART-ft and SixT for all translation pairs. Overall, it underperforms SixT+ (a concurrent work), but ties for si-en, and bests it for my-en (16.5 > 15.3).\footnote{\citet{chenMakingMostMultilingual2022} did not provide is-en results, but their model should support it.}

\paragraph{Unsupervised NMT Results}
We next compare our best unsupervised model, EcXTra-$r_2$ to ~\citet{garcia-etal-2021-harnessing}, the only prior work, to the best of our knowledge, that also trains a single bidirectional unsupervised NMT model. $r_2$ notably achieves a new state-of-the-art for unsupervised en-kk (22.9 > 10.4), and also improves on kk-en (18.2 > 16.4) and si-en (17.8 > 16.2). $r_2$ underperforms for gu-en (13.9 < 16.4) and ne-en (19.7 < 21.7).

Our work is the first to report unsupervised NMT on en-ps, en-is, and en-my.  For an upper bound we cite prior results from supervised NMT systems; these are for reference only (and not even necessarily bidirectional nor multilingual). As expected, $r_2$ underperforms for most tasks. However, $r_2$ notably exceeds supervised results for en-is (25.4 > 23.6), showing the strength of our approach.

\section{Discussion and Analysis}
\label{sec:discuss}

Enabling English-foreign translation in the second stage seems to come at the cost of some foreign-English performance. This may be an instance of the insufficient modeling capacity problem of multilingual NMT models~\cite{zhang2020improving}. Still, $r_2$ improves over $r_1$, while training on entirely synthetic parallel data generated from back-translations in both directions. This finding underscores the effectiveness of successive rounds of back-translation.

The EcXTra-trained model $r_0$ underperforms SixT+~\cite{chenMakingMostMultilingual2022} for foreign-English translations. Because EcXTra is a training approach, we can use SixT+ as a drop-in replacement for $r_0$ for both weight initialization, and for its back-translations. We suspect that training such a combined model would achieve even better English-foreign performance, and leave this to future work.

The EcXTra-trained model $r_2$ underpeforms \citet{garcia-etal-2021-harnessing} for English-Indic translations. This is likely a function of our \textit{m2e-40} dataset having a much lower proportion of Hindi that the dataset of \citet{garcia-etal-2021-harnessing}.\footnote{This is not explicitly specified in their paper, but is clear given their 4 auxiliary languages, vs our 40.} While we take an agnostic view of multilinguality, our training data is by no means writing script-centric; possibly making our model worse at outputting Indic texts. The exceeding en-kk and high en-is scores of $r_2$ provide some evidence for this.

Overall, the $r_2$ achieves competitive unsupervised translation results. Our model supports 3 additional language pairs over prior bidirectional unsupervised translation models, and the EcXTra approach makes it simple to extend to even more translation pairs. We underscore the overall appeal of our approach, in that we can use the zero-shot model to bootstrap back-translations for any unseen language, and train a bidirectional translation system from there.

\subsection{Many-to-English Performance of Unsupervised Models}
\label{ssec:m2e_unmt}
Unlike for the zero-shot $r_0$, the unsupervised $r_2$ has seen text in the text languages, albeit as synthetic parallel sentences with English. A natural question to ask is whether $r_2$ is able to maintain many-to-English performance for non-test languages.

We perform the following experiment to examine this. The models are tasked with \textit{supervised} translation from 4 train languages (zh, hi, tr, ru) to English. $r_0$ and $r_1$ directly see these in their training parallel data, whereas $r_2$ has only indirectly seen them through the prior rounds.

\mtoecheck
The results are shown in Table~\ref{tab:m2echeck}. As was found for the test languages, $r_1$ performs worse than $r_0$. $r_2$ has the same average BLEU across language pairs as $r_1$.  From this short experiment we have shown that the unsupervised models $r_1$ and $r_2$ do retain reasonable Many-to-English performance. We leave future work to investigate mitigation of the forgetting of prior learned tasks, endemic to (almost) all deep learning-based models.
\section{Related Work}
\label{sec:related}
The field of low-resource and zero-resource neural machine translation is an area of continued interest. Below, we describe related works those which follow our data constraint: parallel foreign-English data in auxiliary languages, and monolingual data in unseen languages.

\subsection{Many-to-English zero-shot NMT Models}
\label{related_m2e}
\citet{chen-etal-2021-zero} propose SixT, a fine-tuning method for foreign-English zero-shot NMT. They initialize both the encoder and decoder to XLM-R. They follow a two-stage fine-tuning approach, first only fine-tuning the decoder layers, then continuing training by unfreezing the encoder layers and decoder embeddings.
The model is trained on a parallel corpora in only de-en, and they report zero-shot to-English performance for 10 languages. 

\citet{chenMakingMostMultilingual2022} propose SixT+, which builds upon the authors' prior work, and is trained on a parallel corpus in 6 source languages. This is concurrent to the first submission of our work. They show their model can address zero-shot tasks from NMT to cross-lingual abstractive summarization. This work has the same goal as our first stage of training.\footnote{\citet{chenMakingMostMultilingual2022} does perform a small-scale study on back-translation for translating English-foreign, but these models are neither multilingual nor bidirectional.} The main differences are in our training data (40 vs 6 source languages, 80M vs 120M pairs), and our simpler zero-shot training stage (no unfreezing, no position disentangled encoder).

\subsection{Unsupervised MT Models}

\paragraph{Utilizing Both Parallel and Monolingual Data}
\citet{ko-etal-2021-adapting} propose NMT-Adapt, a method which follows the same data constraints as our work. Their method jointly optimizes four tasks: denoising autoencoder, adversarial training, high-resource translation, and low-resource back-translation -- the latter two of which we also use. However, their work trains individual models for each direction, and furthermore for each model explicitly trains on related high-resource language datasets. This approach is thus more expensive and less adaptable to new languages as ours.

\paragraph{Bidirectional Multilingual NMT}
\citet{garcia-etal-2021-harnessing} train a single model to translate unseen languages to and from English, under the same data constraints as our work. They proceed in 3 stages, each of which uses a mixture of training data and objectives: \textit{MASS}~\cite{song2019mass} for monolingual data, \textit{cross-entropy} for auxiliary parallel data, and both \textit{iterative back-translation}~\cite{hoang2018iterative} and \textit{cross-translation}~\cite{garciaMultilingualViewUnsupervised2020} for synthetic parallel data. This work shares our goal of developing a single bidirectional UNMT model for unseen languages. There are two main differences. First, their aforementioned training scheme is fairly involved. Second, their approach relies on cross-translation, which explicitly ties individual auxiliary languages to unseen languages, limiting their model's cross-lingual generalizability.

\section{Conclusion}
We have described a two-stage training approach for developing a single bidirectional, unsupervised NMT model, which we term EcXTra. The main contribution of EcXTra is in its effective synthesis of techniques from both zero-shot NMT, multilingual fine-tuning, and from unsupervised NMT, back-translation. While prior work also uses similar underlying techniques, they have much more involved training processes, either to consider the bidirectional and zero-shot direction, or introduce additional loss functions (which make training more involved). Furthermore, in this work we have taken an agnostic view towards multilinguality.

We trained a single NMT model through EcXTra, and find that each round of back-translation training further refines bidirectional translation performance. This gives rise to the view of EcXTra as successive rounds of informed initialization into further fine-tuning. The final, unsupervised EcXTra-trained model achieves competitive performance on 7 foreign-English tasks, in both directions. The straightforward nature of EcXTra allows it to be easily extended to new unseen languages.


\section*{Limitations}
\label{sec:limitations}
The notable limitations are the datasets used, the compute required for training, and a want for further ablation studies.

Our training dataset \textit{m2e-40} is a subset of the Many-English dataset~\cite{gowda-etal-2021-many}. This is a collection of various datasets, many of which contain mined parallel sentences. While we have assumed in our paper, like prior work, that these datasets are ``real'' parallel data, they are in fact quite noisy, and contain many low-quality sentence pairs that likely harm downstream system performance~\cite{kreutzerQualityGlanceAudit2022}.

Another potential limitation is that when we select only 2 million samples for each training language pair, instead of using all samples, we limit performance. This is possible, but our work explores a language-agnostic multilingual setting. We refer the interested reader to~\cite{zhangHowRobustNeural2022}, which finds through an empirical study that overall multilingual translation performance is best when languages are balanced.

Our method requires a solid amount of computing resources in order to train the entire NMT system (see details in Appendix~\ref{appsec:setup}). Unlike several other works, we train a single model for all directions, which allows us to be more resource-efficient. However, very recent work has found that even without fine-tuning, multilingual pretrained LMs are able to perform zero-shot translations to and from low-resource languages~\cite{patel2022bidirectional} -- so long as they are given few-shot examples (which can even be synthetic). We suspect such in-context learning based approaches will be soon popular in machine translation, as they have become in many other NLP fields.

We also note that in our work, we evaluated using only BLEU scores. BLEU, of course, is widely-used and understood in the MT community. However, over the decades, researchers have called into question relying solely on BLEU results for MT evaluation. We acknowledge this point, and keep our work as-is given our resource limitations, and given our consistency with prior unsupervised NMT work on reporting results. 

\subsection{Preliminary Ablations}
We understand that ablation studies are useful to ascertain the contribution of various parts of the training approach. Unfortunately, we were unable to pursue this in detail because of resource limitations on our end. Therefore, we enumerate several possible ablations here, and provide preliminary observations from some small-scale experiments:

\paragraph{Model Size} We found the large models for XLM-R and RoBERTa, instead of the base models, significantly increased performance for all language pairs and directions.

\paragraph{Our Dataset vs. Prior Work Datasets}
In the unsupervised and zero-shot NMT literature, because of the variety of task formulations and setups, works do not use consistent datasets for training. This is true for the models we provide reference comparisons to, \citet{chenMakingMostMultilingual2022} and \citet{garcia-etal-2021-harnessing}. These works, like ours, provide comparisons to prior work, with a disclaimer that these results cannot be completely fair. To some extent, the multilinguality agnostic dataset is a key part of the full EcXTra approach. Still, an elucidating ablation experiment could be to train our first stage model using the AUX6 dataset of~\citet{chenMakingMostMultilingual2022}, then run back-translations using the monolingual datasets specified by~\citet{garcia-etal-2021-harnessing}. However, this would require additional computational resources that we unfortunately lack. 

\paragraph{Unidirectional Unsupervised NMT} We found a unidirectional English-foreign second stage model achieves similar BLEU to the bidirectional second stage models.
This suggests that this MT system has no issue with bidirectionality, affirming the findings of~\citet{niu2018bi}.

\paragraph{Bilingual vs. Multilingual NMT Models} We found a second stage model trained to only translate a single bilingual pair, kk-en, performs quite a bit better for those translation directions than a multilingual model. This suggests that the model has difficulty with maintaining performance given all the different translation tasks, especially those with unique scripts such as Burmese and Nepali. 

Training models for individual language pairs (with their own limited vocabularies), and tailoring the datasets specifically to relevant high-resource languages, is one approach as performed by~\cite{ko-etal-2021-adapting}. For example, their ne-en specific model achieves 26.3 BLEU vs. EcXTra's 8.8.\footnote{Still, in the ne-en direction their models achieves only 18.8 BLEU (vs. EcXTra's 19.9) This suggests the multilingual similarities are currently better exploited for to-English translation, than from-English.}. However, this approach is still someone unsatisfying, as our ultimate goal is still to train a single multilingual NMT system. We hope for continued research to close this gap between multilingual and bilingual NMT systems.

\paragraph{Initializing Stage 2 to Stage 1 Model} In this experimental setting, we use the trained stage 1 model only to create English-foreign synthetic parallel data, but initialize to RoBERTa and XLM-R (instead of the stage 1 model). We ran this model for a few epochs, before stopping it because we found the validation BLEU increased very slowly relative to the original stage 2 training. This affirms our earlier claim that the stage 1 model is an informed initialization for the stage 2 model.

\bibliography{anthology,custom}
\bibliographystyle{acl_natbib}

\clearpage
\appendix
\section{Details on Datasets Used}
\label{appsec:data}
Here, we expand upon Section~\ref{sec:data} and provide further detail on the datasets used in this paper.

\subsection{Zero-Shot NMT Datasets}
\paragraph{Test}
We consider translation of 7 low-resource languages, which come from 6 language families. We draw these test sets from publicly available datasets from WMT21\footnote{\url{https://www.statmt.org/wmt21/index.html}},  FLoRes v1\footnote{\url{https://github.com/facebookresearch/flores/tree/main/floresv1}}, and WAT21\footnote{\url{http://lotus.kuee.kyoto-u.ac.jp/WAT/my-en-data/}}. Where possible, we use the same test sets as specified by prior unsupervised NMT work. 

\paragraph{Training} 
Our first stage model is trained on a parallel dataset we term \textit{m2e-40}. This is a subset of the Many-English\footnote{\url{http://rtg.isi.edu/many-eng/data-v1.html}} dataset~\cite{gowda-etal-2021-many}, which itself is a collection of other publicly available datasets. Of the 500 language pairs in this dataset, we choose the 40 languages with the most parallel sentences\footnote{The motivation for choosing 40 languages is largely because of resource limitations on our end. Ideally, we would have liked to train on all languages with 1M+ sentence pairs.}. This criterion contrasts with prior work~\cite{siddhant2022towards, chenMakingMostMultilingual2022}, which specifically select language pairs based on coverage and/or similarity to the unseen test languages. Table~\ref{tab:train_langs} shows more information for the training languages.

Prior work has handled the imbalance in auxiliary language pairs through temperature sampling~\cite{Devlin2019BERTPO}. Essentially, this is a simple trick to up-sample high-resource languages and down-sample low-resource once. In our work we take the even simpler trick of equally sampling 2 million sentences from each training language. This follows the finding of~\citet{zhangHowRobustNeural2022} that more equal sampling of languages results in the relatively best multilingual performance.

The Many-English dataset is provided as pre-tokenized and pre-processed. For our use-case, we are fine-tuning the encoder of XLM-R, which was pretrained on untokenized text. Therefore, we detokenize both the English and the foreign sides of our subset using \texttt{sacremoses}\footnote{\url{https://github.com/alvations/sacremoses}}.

\paragraph{Validation}
The validation data comes from the development tarball of WMT19\footnote{\url{http://data.statmt.org/wmt19/translation-task/dev.tgz}}. Of the 40 training languages, 15 of them are found in this tarball. As some translation directions appear multiple times (e.g. fr-en), we choose just 1 per task. Table~\ref{tab:val_langs} shows more information. For the supervised NMT experiment of Section~\ref{ssec:m2e_unmt}, we utilize the same development datasets for the languages \{zh, hi, tr, ru\}.

\subsection{Unsupervised NMT Datasets}

\paragraph{Training} 
We use several monolingual datasets for training our unsupervised NMT model. For the 7 test languages we draw from Common Crawl\footnote{\url{https://data.statmt.org/ngrams/}} for \{kk, gu, is\} and CC-100\footnote{\url{https://data.statmt.org/cc-100/}} for \{my, ps, ne, si\}.

We take the first 4M sentences of each monolingual dataset--except for Burmese (my), which has only 2M sentences. We then filter out duplicated lines, and empty lines. We thus have 26M test language sentences.

For the English-to-many direction, we require monolingual English data, which we draw from News crawl\footnote{\url{https://data.statmt.org/news-crawl/en/}}. As above, we take the first 4M sentences, then filter out duplicated and empty lines. The English monolingual sentences are then translated in the 7 languages, resulting in 7 * 4M = 28M synthetic sentence pairs total.

\paragraph{Validation}
For each round of back-translation training, we use datasets in 14 directions -- from/to the 7 translation directions. We withhold the first 250 sentence pairs of each translation direction (14 directions, so 3500 pairs total) to serve as validation. The early stopping criteria is standard BLEU. We tried as an alternative the round-trip BLEU proposed by~\citet{lampleUnsupervisedMachineTranslation2018}, but found this made little difference in final evaluation results.

\section{Modeling and Training Setup}
\label{appsec:setup}
Our research was pursued in a resource-limited setting. 
For training, we used 4 NVIDIA RTX A6000 GPUs (48GB vRAM each). 
For inference, we used the above, and additionally had access to 16 NVIDIA GeForce RTX 2080 Ti GPUs (11GB vRAM each).

Given the above resource-limited training and inference setup, we provide some rough estimates of runtime. Training a stage 1 model takes about 1 week. Training a stage 2 model takes about 6 weeks, given the steps: a) run xx->en back-translations on 26m sentences (2 weeks), b) train the round 1 model (1 week), c) run en->xx back-translations on 28M sentences (2 week), d) train the round 2 model (1 week). Given more standard GPU resources, we would expect at least a 3-4x speedup in the whole training process.

We use the \texttt{transformers} package~\cite{wolf2020transformers} as the backbone for our modeling work. Specifically, we use it to load pretrained model weights and tokenizers. The rest of the code is implemented in PyTorch~\cite{paszke2019pytorch}.

\paragraph{Hyperparameters}
The most up-to-date version of the hyperparameters can be found in the repository.\footnote{\url{https://github.com/manestay/EcXTra/}} For training, the batch size = 20000 for round 0, and 11500 for rounds 1 and 2. We use an Adam optimizer, with learning rate = 1e-3, and warmup steps = 12500. The learning rate decay schedule is based on the inverse square root of the update number. The dropout probability = 0.1, and the random mask probability = 0.4. For inference, the batch size = 1500, and beam size = 5.

\section{Start Tokens to Indicate Target Language}
\label{ssec:start_tok}
Following~\citet{johnsonGoogleMultilingualNeural2017}, we add special start tokens to each source sentence, to indicate the desired target language. This only applies to stage 2, because stage 1 always targets English. The default implementation directly adds these tokens, of the form \texttt{<2xx>} to the target vocabulary. Our setting requires adapting the implementation because as we have frozen the target embeddings (and source embeddings), we cannot increase the vocabulary size. We therefore indicate the target language with a two-token sequence, which consists of the usual start token \texttt{<s>}, and another token \texttt{TOK\textsubscript{i}} drawn from the long tail of the vocabulary. The model then must learn that \texttt{<s> + TOK\textsubscript{i}} means to translate to a given language.

To be concrete, we use XLM-R tokenization, which consists of 250,002 SentencePiece tokens. For this paper, in which the model supports 8 languages, we arbitrary select indices $202201$ to $202208$, and assign each to a language.


\section{How Zero-Resource is Zero-Resource?}
In this work, we have defined zero-resource as the setting in which no parallel sentences are available for a language pair of interest. This definition follows the general usage in the field. To be exactly precise, though, the pretrained multilingual model used, XLM-RoBERTa, has indeed seen monolingual text in each of the 7 low-resource languages.


\begin{table*}[ht]
    \centering
    \begin{tabular}{llll|ll}
    \toprule
    Code & Language & Family & Script & Source & \# Pairs \\
    \midrule
    kk & Kazakh & Turkic & Cyrillic & newstest2019 & 1000\\ 
    gu & Gujarati & Indic & Gujarati & newstest2019 & 1016 \\ 
    si & Sinhala & Indic & Sinhala & FLoRes v1 & 2905 \\
    ne & Nepali & Indic & Devanagari & FLoRes v1 & 2924 \\
    ps & Pashto & Iranian & Arabic & newstest2020 & 2719 \\
    is & Icelandic & Germanic & Latin & newstest21 & 1000 \\
    my & Burmese & Burmese-Lolo & Burmese & WAT21 & 1018 \\
    \bottomrule
    \end{tabular}
    \caption{Information for the \textbf{test} languages, and the foreign-English datasets used. The columns are, from left to right, the ISO 639-1 language code, the name of the language, the language family at the Genus level, the data source, and the number of sentence pairs.}
    \label{tab:test_langs}
\end{table*}

\begin{table}[ht]
    \centering
    \begin{tabular}{ll|ll}
    \toprule
    Code & Language & Code & Language \\
    \midrule
    tr & Turkish &    hu & Hungarian \\
    sr & Serbian &    sl & Slovenian \\
    fr & French &    vi & Vietnamese \\
    he & Hebrew &    et & Estonian \\
    ru & Russian &    sk & Slovak \\
    ar & Arabic &    ja & Japanese \\
    zh & Chinese &    lt & Lithuanian \\
    bs & Bosnian &    lv & Latvian \\
    nl & Dutch &    uk & Ukrainian \\
    de & German &    th & Thai \\
    pt & Portuguese &    cs & Czech \\
    no & Norwegian &    ko & Korean \\
    it & Italian &    id & Indonesian \\
    es & Spanish &    ca & Catalan \\
    pl & Polish &    mt & Maltese \\
    fi & Finnish &    ro & Romanian \\
    fa & Persian &    bg & Bulgarian \\
    sv & Swedish &    hr & Croatian \\
    da & Danish &    hi & Hindi \\
    el & Greek &    eu & Basque \\
    \bottomrule
    \end{tabular}
    \caption{Information for the \textbf{train} languages.  The columns are, from left to right, the ISO 639-1 language code, and the name of the language.}
    \label{tab:train_langs}
\end{table}

\begin{table}[htb]
    \centering
    \begin{tabular}{ll|ll}
    \toprule
    Code & Language & Source & \# Pairs \\
    \midrule
    tr & Turkish & newsdev2016 & 1001 \\
    fr & French & newstest2009 & 2525 \\ 
    ru & Russian & newstest2012 & 3003 \\
    zh & Chinese & newsdev2017 & 2002 \\
    de & German & newstest2009 & 2525 \\
    it & Italian & newstest2009 & 2525 \\
    es & Spanish & newstest2009 & 2525 \\ 
    fi & Finnish & newsdev2015 & 1500 \\
    hu & Hungarian & newstest2009 & 2525 \\
    et & Estonian & newsdev2017 & 2000 \\
    lt & Lithuanian & newsdev2019 & 2000 \\
    lv & Latvian & newsdev 2017 & 2003 \\
    cs & Czech & newstest2009 & 2525 \\
    ro & Romanian & newsdev2016 & 1999 \\
    hi & Hindi & newsdev2014 & 520 \\
    \bottomrule
    \end{tabular}
    \caption{Information on the \textbf{validation} languages, and the foreign-English datasets used. The columns are, from left to right, the ISO 639-1 language code, the name of the language, the source (from WMT development set), and the number of sentence pairs.}
    \label{tab:val_langs}
\end{table}

\section{Sample Output}
Sample output for the EcXTra NMT models are shown in Tables~\ref{tab:sample_transl1} and~\ref{tab:sample_transl2}.
\begin{table*}[ht]
\centering
\begin{tabular}{@{}lp{13cm}@{}}
\toprule
Model & \begin{tabular}[c]{@{}l@{}}Translation (kk-en) \end{tabular} \\ \midrule
Reference & The first medal place was given to Dastan Aitbay from Kyzylorda and his project on "Safe Headphones" Innovative headphones". \\ \midrule
EcXTra-$r_0$ & The winning first place was won by Dastan Aitbay's innovative earpiece "Safe headphones" from the city of Kyushu. \\ \midrule
EcXTra-$r_1$ & First place was won by Dastan Attbay of the city of Kyrgyzlord "Innovative earphones "Safe headphones." \\ \midrule
EcXTra-$r_2$ & The cool first place was won by Dastan Aitbay, from the city of Kyrgyzstan, the "Inventive earcap Safe Headphones." \\ \bottomrule
\end{tabular}
\caption{Sample kk-en unsupervised translations for the input: \foreignlanguage{russian}{
Жүлделі бірінші орынды Қызылорда қала­сынан Дастан Айтбайдың "Инновациялық құлақ­қап "Safe headphones" жобасы жеңіп алды.}}
\label{tab:sample_transl1}
\end{table*}

\begin{table*}[ht]
\centering
\begin{tabular}{@{}lp{13cm}@{}}
\toprule
Model & \begin{tabular}[c]{@{}l@{}}Translation (en-is) \end{tabular} \\ \midrule
Reference & Markmiðið er að fegra svæðið og leyfa mósaíkverki Gerðar Helgadóttur á Tollhúsinu að njóta sýn betur. \\ \midrule
EcXTra-$r_0$ & N/A \\ \midrule
EcXTra-$r_1$ & Markmið er að fagna svæðið og gera mosaík Gerður Helgadóttir á Tollhúsinu áberandi. \\ \midrule
EcXTra-$r_2$ & Tilgangurinn er að fallega svæðið og gera mosamynd Gerður Helgadóttir á tollhúsinu meira áberandi. \\ \bottomrule
\end{tabular}
\caption{Sample en-is unsupervised translations for the input: The aim is to beautify the area and make Gerður Helgadóttir's mosaic on the Customs House more prominent.}
\label{tab:sample_transl2}
\end{table*}

\end{document}